\pdfoutput=1

\documentclass[11pt]{article}

\usepackage{acl}
\usepackage{graphicx}
\usepackage{soul}
\usepackage{float}

\usepackage{times}
\usepackage{latexsym}
\usepackage{amsmath}
\usepackage{amssymb}

\usepackage{hyperref}       
\usepackage{url}             
\usepackage{booktabs}       
\usepackage{amsfonts}       
\usepackage{nicefrac}       
\usepackage{microtype}      
\usepackage{lipsum}		
\usepackage{graphicx}
\usepackage{natbib}
\usepackage{caption}
\usepackage{subcaption}
\usepackage{doi}
\usepackage{xcolor}

\usepackage[T1]{fontenc}

\usepackage[utf8]{inputenc}

\usepackage{microtype}

\title{Relational Extraction on Wikipedia Tables using Convolutional and Memory Networks}

\author{Arif Shahriar$^*$ , {\bf Rohan Saha$^*$},  {\bf Denilson Barbosa} \\
        Department of Computing Science \\ University of Alberta \\
        \{ashahri1, rsaha, denilson\}@ualberta.ca}

\begin{document}

\maketitle
\def\thefootnote{*}\footnotetext{Equal Contribution}\def\thefootnote{\arabic{footnote}} 
\begin{abstract}
Relation extraction (RE) is the task of extracting relations between entities in text. Most RE methods extract relations from free-form running text and leave out other rich data sources, such as tables. We explore RE from the perspective of applying neural methods on tabularly organized data. We introduce a new model consisting of Convolutional Neural Network (CNN) and Bidirectional-Long Short Term Memory (BiLSTM) network to encode entities and learn dependencies among them, respectively. We evaluate our model on a large and recent dataset and compare results with previous neural methods. Experimental results show that our model consistently outperforms the previous model for the task of relation extraction on tabular data. We perform comprehensive error analyses and ablation study to show the contribution of various components of our model. Finally, we discuss the usefulness and trade-offs of our approach, and provide suggestions for fostering further research. 
\end{abstract}

\section{Introduction}


Knowledge graphs (KG) are important lexical resources for various applications involving natural language, such as web searches, question answering, etc. However, KGs quickly become incomplete as the world changes. Therefore, adding new facts to a KG is crucial for maintaining its relevance. Relation extraction (RE) is the task of extracting relations between two entities in a piece of text. RE has been widely used as a way of KG completion. 
Although there is a plethora of work in relation extraction, most methods process continuous free-form text (e.g., complete sentences) mentioning entities, leaving out other important data sources such as tables.  

\par Unlike previous works that used neural networks on continuous text \cite{Lin2016, Zheng2017, Su2018, xing-luo-2019-distant, lee2019semantic, zeng-etal-2015-distant}, we focus on extracting relations from tabular data. 
We use a neural model for our analysis as neural methods have been shown to outperform traditional RE approaches that require feature engineering;  \citet{Wang_2022} give a recent review of neural methods in relation extraction. 
The model extracts relations between a pair of entities in different columns inside a table and, for encyclopedic and biographical articles, between the subject of the article and an entity in a table inside that article. 
The model uses a combination of convolutions and memory networks to automatically extract useful features and model dependencies among features, respectively. We show that our approach can consistently outperform and makes fewer errors than a previous model. \par

Our main contributions are as follows.
\begin{enumerate}
    \item We outperform a state-of-the-art neural model for extracting relations from table data. 
    \item We perform a comprehensive error analysis to highlight the cost of model parameters for a comparable performance gain.
    \item Analyze the model performance for individual relations and investigate the strengths and limitations of the proposed method.
\end{enumerate}

All of our code is provided in this repository: \url{https://github.com/simpleParadox/RE_656}

\section{Related Work}
\label{sec:related_work}

Most prior works have mainly focused on sentence-level RE where deep neural networks have been used to assign relations for a pair of entities \citep{Lin2016, zeng-etal-2015-distant, Zheng2017, xing-luo-2019-distant, lee2019semantic}. Recent works have also moved the research direction from sentence level to document level RE to utilize richer information in documents and perform relation extraction across sentences. For document-level relation extraction, recent works have also used techniques such as constructing a document-level graph using dependency trees, coreference information, rule-based heuristics, and Graph Convolutional Networks (GCN) \citep{sahu-etal-2019-inter, christopoulou-etal-2019-connecting, nan-etal-2020-reasoning} for reasoning and predicting relations. As evident, RE from continuous text is explored widely, but only a few papers have addressed the task of RE from data that is non-free form, such as data organized into tables \citep{10.1145/3340531.3412164, munoz2014using}.

We need features that accurately describe the input data for the relation classification task. These features can be manually created or automatically learned from the input. \citeauthor{munoz2014using} used manual feature-engineering techniques and traditional machine-learning models to extract relations in the form of Resource Description Framework (RDF) triples from tabular data. Although their method achieved an F1-score of 79.40\%, it requires complicated manual feature engineering. On the contrary, most recent works overcome the task of manual feature engineering using end-to-end deep learning techniques, and we use a similar motivation to use neural models for automating feature extraction for relation classification.


The most notable work related to ours is the one by \citeauthor{10.1145/3340531.3412164}, looking at extracting relations from a given pair of entities in Wikipedia tables. They used embeddings from BERT \cite{Devlin2019} and a simple neural network with 1 LSTM unit to classify relations. Although a highly effective approach, we found the method to be over-simplistic to properly capture many relations. 
We show that a more sophisticated model involving convolutions and bidirectional-LSTM may be a better approach for the task of classifying relations for entity pairs from tabular data.

The choice of convolution networks here is justified by the many previous works showing that CNNs perform significantly better than traditional feature-based methods for relation extraction. Each instance in our data is composed of multiple components such as table headers, table caption, section title containing the table etc. A CNN will automatically learn the useful features, and then finally, max-pooling merges them to perform predictions globally. Previous works such as \citet{zeng-etal-2015-distant} introduced the convolutional architecture with piecewise max pooling (PCNN) to capture structural information between entities and adopted multi-instance learning into PCNN for a dataset that was built using distant supervision \cite{mintz-etal-2009-distant}. They divided the input sentence into three segments and applied a max-pooling operation on each segment instead of the entire sentence. Secondly, \citet{Lin2016} used a CNN model for an RE task with sentence-level attention for multi-instance learning, where the model used informative sentences and de-emphasized noisy samples. Finally, \citet{xing-luo-2019-distant} proposed a novel framework that uses separate head-tail convolution and pooling to encode input sentences and classified relations from coarse to fine to filter out negative instances. Therefore, the papers mentioned above have shown the effectiveness of CNN for automatically learning features from sentences.\par

Hybrid neural models have also been shown to perform well in RE tasks. \citet{Zheng2017} introduced a hybrid neural network (NN) that consists of a bidirectional encoder-decoder LSTM module (BILSTM-ED) for named entity recognition and a CNN module for relation classification. Initially, they used BILSTM-ED to capture context and then fed obtained contextual information to the CNN module to improve relation classification. Furthermore, an encoder-decoder-based CNN+LSTM approach has been presented by \citet{Su2018} for distant supervised RE. Their CNN encoder captured sentence features from a bag of sentences and merged them into a bag representation, and the LSTM decoder predicted relations sequentially by modelling relations' dependencies. As hybrid networks have shown their utility for the RE task, we utilize a hybrid architecture for relation classification from tabular data.
\par
The utility of BiLSTM is also evident in tackling the task of RE. \citet{lee2019semantic} proposed an end-to-end recurrent neural model incorporating an entity-aware attention mechanism with latent entity typing. They applied BiLSTM to build recurrent neural architecture to encode the context of the sentence. We also include a BiLSTM as a component of our model since it has been shown to perform well on RE tasks by modelling contextual information and leveraging long-term dependencies.

\section{Methods}
\label{sec:methods}

Here, we describe our task and our model in detail. 

\begin{figure*}[t]
    \centering
    \includegraphics[width=\textwidth]{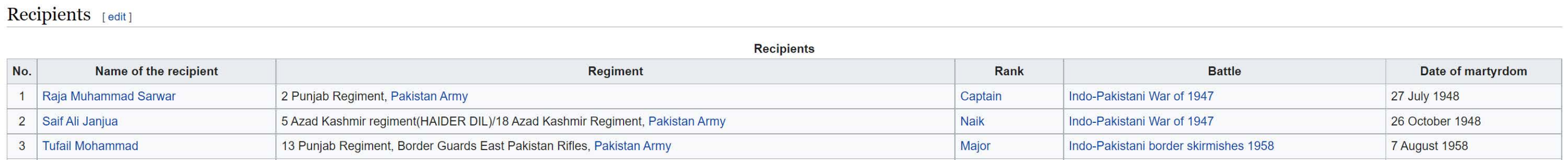}
    \caption{Table from the Wikipedia article for ``Nishan-e-Haider''. The head and tail entities can be the cell values in the same row but different columns, or, the article title (Nishan-e-Haider) and any of the cell values of the table.}
    \label{fig:wikipedia_table}
\end{figure*}

\subsection{Task}

The task is to extract relations between a pair of entities in which one or both appear inside a table. This task has been studied in the context of Wikipedia, so we use that encyclopedia in our discussion for clarity.  Recall that each Wikipedia article is about a single entity, which is called the (entity) \textit{subject} of that article. Our task is then to find relations either between a pair of entities appearing on the same row (but different columns) of a table inside an article, or between an entity appearing inside a table and the subject entity of the article.

For example, consider a table from the Wikipedia article ``Nishan-e-Haider'' shown in Figure \ref{fig:wikipedia_table}. Each entity under the ``name of the recipient” column (``Raja Muhammad Sarwar") is a recipient of the award ``Nishan-e-Haider''\footnote{\url{https://en.wikipedia.org/wiki/Nishan-e-Haider}}. 
Therefore, the article subject has a relation (award-nominee) with the recipient entity in the table cell. Furthermore, elements of the article besides table cell values, like a column header (``Name of the recipient”), table section title, and caption (``Recipients") provide additional contextual information to identify the relation ``award-nominee” between corresponding entity pairs.

\subsection{Embeddings} 
\label{sec:embeddings}
Before training our model, we obtain vector representations of our input. For each table in the dataset, we tokenize the table cell values representing the subject and object entities.  We also use contextual information from the table, including the title of the section containing the table and table headers and captions (if present).
In addition, we use the subject and object column indices to obtain related entity pairs for a table row. We do not use the table section paragraphs as \citet{10.1145/3340531.3412164} found no gain in performance by including them. 


We concatenate the entity pairs and the contextual information to obtain a training sample for a given relation. We then preprocess the sample and remove all non-alphanumeric characters (e.g. $<$SEP$>$ token, brackets []) using Python’s \texttt{regex} module. Then we use the pretrained BERT tokenizer\footnote{\url{https://github.com/google-research/bert/blob/master/tokenization.py}} based on the WordPiece to tokenize the inputs. To obtain a vector representation of the concatenated input, we use HuggingFace's implementation of BERT (base\_uncased) \cite{Devlin2019} pretrained on Wikipedia and BookCorpus and trained in an uncased fashion. We set the max length of the input to consist of 80 tokens, compared to the previous work by \citet{10.1145/3340531.3412164}, which used 50 tokens. We retrieve a 768-dimensional word embedding for each token and then concatenate all the embeddings to represent the sample. We used BERT embeddings because they have been shown to perform well in various NLP tasks \cite{baldini-soares-etal-2019-matching, wang2019fine, nan-etal-2020-reasoning, 10.1007/978-3-030-47426-3_16}.
Moreover, we use contextual clues for tables for relation extraction which justifies the use of contextual word embeddings.


\subsection{Convolutional Neural Network}   
\label{sec:cnn}
As customary \cite{Lin2016, xing-luo-2019-distant, zeng-etal-2015-distant}, we fed the instance embeddings to a convolutional layer as it is capable of merging all the local features in input sentences. Since we are considering all surrounding information around the table, important information can appear anywhere in the input sentence. Therefore, it is necessary to leverage all local features and contextual clues in input samples. Convolution involves a dot product of the weight matrix with every k grams in the sequence S to obtain latent feature\(\ C^{(i)}\ \), which is shown in equation \ref{eq1}.\(\ W_c^{(i)}\in\mathbb{R}^{k\ \times\ d}\ \) indicates $i_{th}$ convolutional filter, k indicates context window size of the learnable filter and \(\ b^{(i)}\ \) indicates bias term. 
To ensure input dimensions are consistent, we padded with zeros evenly to the left and right of the input sequence. Moreover, we employed 8 filters in the convolution process to learn different features. We applied the ReLU non-linear activation to the output for incorporating non-linearity.

\begin{equation}\label{eq1}
   \ C^{(i)}\ =\ W_c^{(i)}\times\ S_{l:l+k-1}\ +\ b^{(i)}\
\end{equation}

Finally, we used max-pooling to preserve the most prominent features derived from each filter, which is defined in the following equation. The max-pooling operation combines all local features to obtain a fixed-size representation of each input sentence.
\begin{equation}\label{eq2}
    C^{(i)}_{max} = max\{C^{(i)}\}
\end{equation}

\subsection{Long-Short-Term-Memory Network}
\label{sec:lstm}
We have used bidirectional long short-term memory networks (BiLSTM) because both earlier and later information can be considered for sequentially modeling contextual information in forward and reverse order. Moreover, LSTM models were successfully applied for relation extraction tasks \cite{Su2018, lee2019semantic} as it uses memory blocks to capture long-term temporal dependencies. \citet{erin-etal-2019} also achieved high performance by using LSTMs to predict relations between pairs of entities in Wikipedia tables. Inspired by their work, we have experimented with BiLSTM to observe any performance increment.

We use BiLSTM to capture interactions among hidden representations obtained from the pooling layer. So, the input to the BiLSTM layer is a sequence obtained from the previous layer  \(C_{max} = \{c_{1}, c_{2}, …, c_{n}\}\). Here, $n$ indicates half of the maximum token length preserved after downsampling the convolutional output representation using the max-pooling operation.
\begin{equation}\label{eq3}
   \overrightarrow{h_{t}} = ForwardLSTM(c_{t}, h_{t-1})
\end{equation}
\begin{equation}\label{eq4}
    \overleftarrow{h_{t}} = Backward LSTM(c_{t}, h_{t-1})
\end{equation}
\begin{equation}\label{eq5}
    x_{t} = [\overrightarrow{h_{t}};\overleftarrow{h_{t}}]
\end{equation}

The BiLSTM consists of two sub-LSTM networks: a forward LSTM and a backward LSTM for modeling dependencies in forward and backward order, respectively. \(\overrightarrow{\text{$h_{t}$}}\) and \(\overleftarrow{\text{$h_{t}$}}\) are the computed outputs at the $t^{\text{th}}$ time step from the forward and backward LSTM. Then, we concatenate hidden states \(\overrightarrow{\text{$h_{t}$}}\) and \(\overleftarrow{\text{$h_{t}$}}\) to obtain the final hidden representation  \(h_{t}\). 

\subsection{Dropout}
We use dropout at the BiLSTM layer for regularization to prevent overfitting. Dropout randomly \textit{turns-off} a fraction of hidden units during the forward pass. It ensures that hidden units can identify features independent of each other rather than showing co-adaption and enable the model to learn a more general representation.

\begin{figure*}[!htb]
    \centering
    \includegraphics[width=\textwidth]{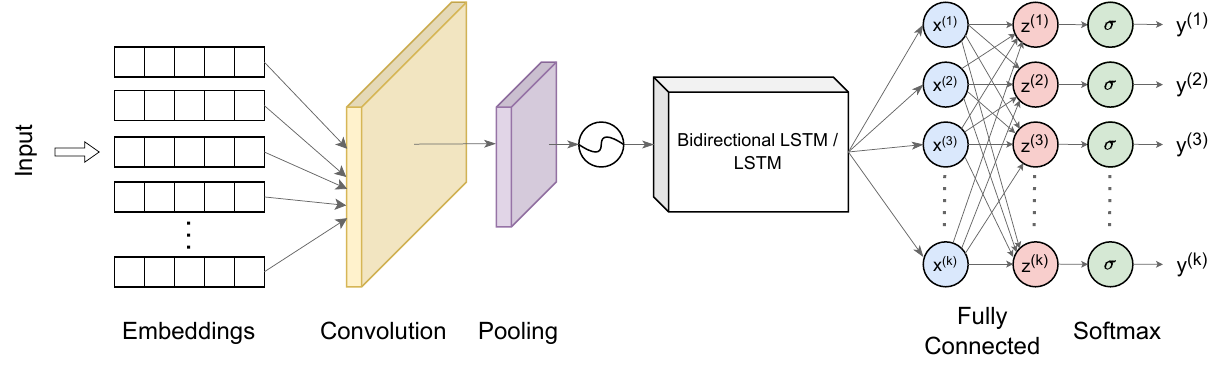}
    \caption{\textbf{Proposed neural architecture.} Concatenated BERT embeddings are passed through a CNN layer, max-pooling layer followed by a ReLU activation function. An LSTM / biLSTM block is used to learn dependencies which is followed by a softmax activation to obtain probabilities for each relation label.}
    \label{fig:arch}
\end{figure*}

\subsection{Classification Layer}
\label{sec:classification}
We feed the output of the LSTM/BiLSTM layer into a fully connected layer. We then take the output of the fully connected layer and apply a softmax function to obtain the probability for each class.
\begin{align*}
    z_k &= W \times X \\
    \hat{y} &= softmax(z_k)
\end{align*}
where X is the output of the LSTM/BiLSTM layer. We show the architecture of our proposed model in Figure \ref{fig:arch}.

\section{Experiments}
\label{sec:experiments}
\subsection{Dataset}
\label{sec:dataset}

We use the data from \citet{erin-etal-2019} in all of our experiemnts. The dataset contains individual JSON files for each relation. These JSON files were obtained from a Wikidata dump from March 2019. We used subject and object column indexes present in the dataset to retrieve the subject and object entity pairs from Wikipedia articles. These subject and object entities indicate related entity pairs in the same row of table or article subject and associated table cell value. Moreover, the dataset also includes table information like the title of the table section, table caption and headers, and table section paragraph. To the best of our knowledge, this is the most recent and the largest dataset created specifically for the task of RE on tabular data.

The dataset was annotated using distant supervision by aligning Freebase entities with mentions of pairs of entities appearing in the table row or article subject and table cell value. The dataset contains 217,834 tables and 29 relations (28 relation types and one \textit{none} relation). The dataset is highly imbalanced, with some relation classes having less than 500 examples. This results in a long-tailed dataset. We do not remove these long-tailed relations.


\subsection{Model Training and Evaluation}
\label{sec:model_train_and_eval}
To train and evaluate our model, we split the dataset into train and test splits. We follow the configurations used by \citet{10.1145/3340531.3412164}, where 40\% of the data was used for training the model, 40\% for validation (for hyperparameter tuning), and 20\% for testing. We use five seeds to obtain train, validation, and test splits and report our results which is the average over the five seeds. We use sparse categorical cross-entropy loss\footnote{\url{https://www.tensorflow.org/api_docs/python/tf/keras/losses/SparseCategoricalCrossentropy}} to train the model.
We used one Nvidia A100 GPU (40GB Memory) for model training.

\subsection{Comparison with Baseline Model}
\label{sec:fair_comparison}
We use the neural relation extraction model proposed by \citet{10.1145/3340531.3412164}, consisting of a single LSTM unit, as the baseline. In order to have a fair comparison with the model introduced by \citeauthor{10.1145/3340531.3412164}, we use F1 and accuracy to measure the performance of our model. We trained the model for forty epochs (as suggested by \citeauthor{10.1145/3340531.3412164}). \par We summarize the number of training parameters of our model and compare it to that of the baseline in Table \ref{tab:parameters_comparison}. We also performed an ablation study where we removed the convolutional layer and investigated the performance of the task for the BiLSTM model only. We show the differences between the hyperparameters of our model and the baseline model in Table \ref{tab:hyperparameters}.

\begin{table}[!htb]
    \centering
    \begin{tabular}{|c|c|c|}
    \hline
        \textbf{Model} & \textbf{Parameters}  \\ \hline 
        \hline
        \citet{10.1145/3340531.3412164} & 4,559\\ \hline
        CNN-LSTM (ours) & 40,581\\ \hline
        CNN-BiLSTM (ours) & 50,405\\ \hline
        BiLSTM (8 units) & 86,877 \\ \hline
    \end{tabular}
    \caption{Comparison of trainable model parameters for baseline \cite{10.1145/3340531.3412164}, our proposed model, and the BiLSTM only model which we use for comparison with our proposed model.}
    \label{tab:parameters_comparison}
\end{table}

\begin{table}[!htb]
\renewcommand{\arraystretch}{1}
    \centering
    \begin{tabular}{|c|c|c|}
    \hline
        \textbf{Hyperparameter} & \textbf{Ours} & \textbf{Baseline} \\ \hline 
        \hline
        CNN Filters & 8 & None\\ \hline
        LSTM/BiLSTM units & 8 & 1\\ \hline
        Batch Size & 16 & 16\\ \hline
        Optimizer & Adam & RMSProp\\ \hline
        Max token length & 80 & 50\\ \hline
        Learning rate & 2e-5 & 0.001\\ \hline
        LSTM/BiLSTM Dropout & 0.2 & None\\\hline 
    \end{tabular}
    \caption{Comparison of hyperparameters between the baseline model and our proposed model.}
    \label{tab:hyperparameters}
\end{table}

\section{Results}
\label{sec:results}

We show the results in Table \ref{tab:results_table}. For relation extraction on tabular data, the previous best model was proposed by \citet{10.1145/3340531.3412164}. Although the performance of the baseline model is significantly high, it may benefit from leveraging automated feature extraction methods, such as using a CNN to extract features. We also add more LSTM units to increase the learning capability of the model. We refer to the upgraded model as CNN+LSTM or CNN+BiLSTM (based on whether we use LSTM or BiLSTM). As we see in Table \ref{tab:results_table}, both CNN+LSTM and CNN+BiLSTM outperform the baseline model and are the current state-of-the-art model for relation extraction on tabular data. The accuracy of the CNN+LSTM model is 5.57\% points higher, and the accuracy of the CNN+BiLSTM model is 5.8\% points higher than the baseline. A higher accuracy will result in more accurately assigning a relation class to an entity pair.\par

We believe that our model performed better because we used 8 BiLSTM units for capturing context and learning dependencies, and 8 CNN filters as a feature extractor. In contrast, \citet{10.1145/3340531.3412164} used only a single LSTM unit for modeling dependencies among input tokens. In comparison to the baseline method that used a maximum token length of 50, we used a maximum token length of 80 to capture more information for each instance. Furthermore, we use dropout that benefits the model, preventing overfitting and ensuring generalizability.

Interestingly, our model was not able to outperform the baseline in terms of F1 score but was still able to provide comparable performance of around 92.46\%. Although a model with better performance will lead to improvements in downstream tasks, for applications such as building knowledge graphs, the performance achieved by our model is sufficient.\par



\subsection{Ablation Study}
\label{sec:ablation_study}
To understand the effectiveness of the convolution layer, we perform an ablation study. We perform the relation extraction on the dataset without using the CNN module, which we refer to as the BilSTM-only model (with 8 units). The number of training parameters is shown in Table \ref{tab:parameters_comparison}. \par
Interestingly, removing the CNN module improves the performance on the task by 6.19\% points more than the baseline. This improvement is likely due to the increase in the number of trainable parameters to over twice that of the CNN+LSTM model. This increase in the number of trainable parameters also leads to a more complex model. Such a result reinforces the prevalent idea that increasing the number of parameters is helpful for the model to learn information from the data. However, this comes at the cost of requiring more computing resources.

\begin{table}[]
    \centering
    \begin{tabular}{|c|c|c|}
    \hline
        \textbf{Model} & \textbf{Accuracy} & \textbf{F1} \\ \hline 
        \hline
        Baseline & 92\% & \textbf{95\%} \\ \hline
        CNN-LSTM &  \textbf{97.57\%} & 91.44\% \\ \hline
        CNN-BiLSTM & \textbf{97.80\%} & 92.46\% \\ \hline
        BiLSTM-only (8 units)  & \textbf{98.19\%} & 94.35\% \\ \hline
    \end{tabular}
    \caption{Performance measures of our approach compared to previous model.}
    \label{tab:results_table}
\end{table}

\begin{figure*}[t]
    \centering
    \includegraphics[width=\textwidth]{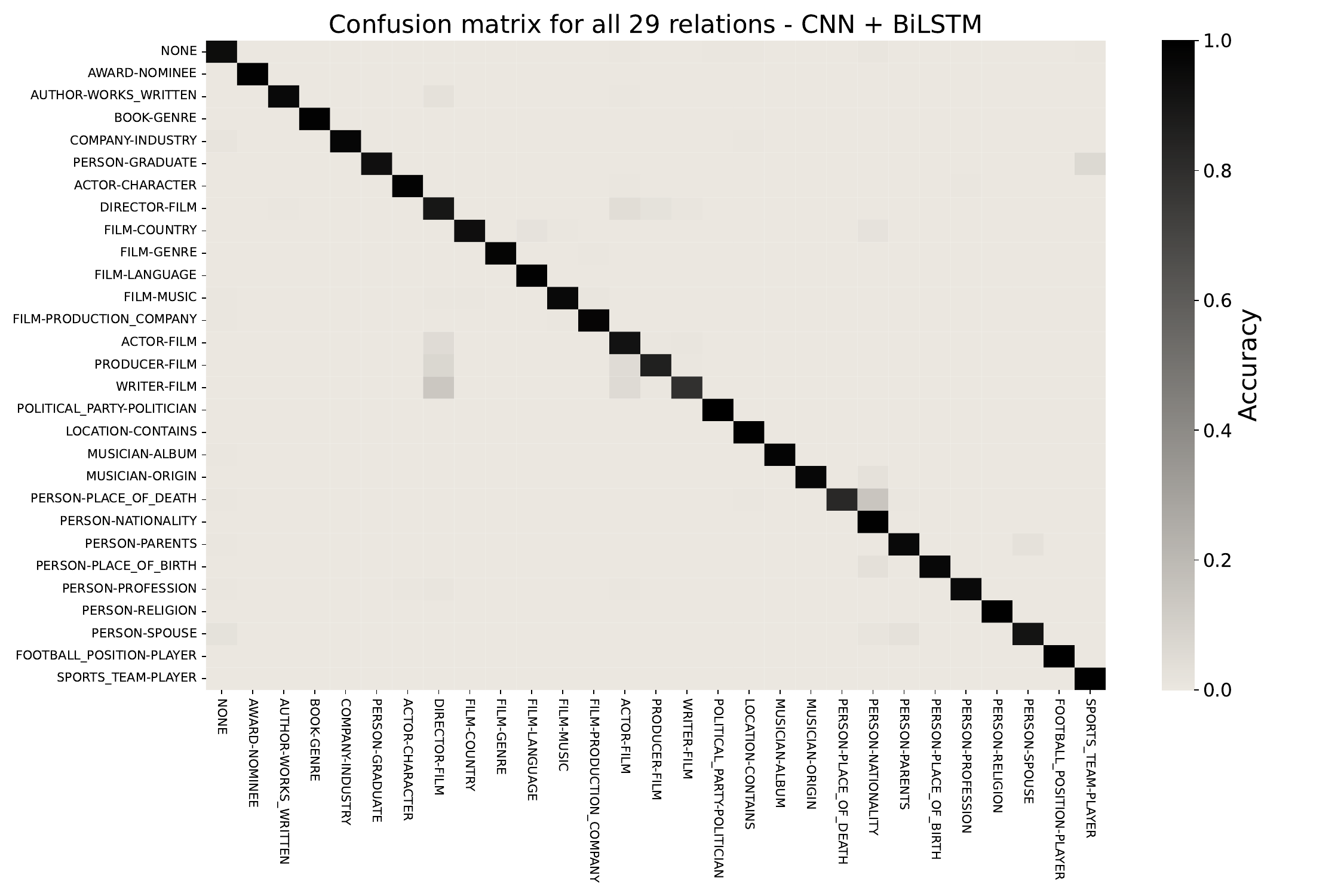}
    \caption{Confusion Matrix for CNN+BiLSTM. The y-axis are the predicted relation labels, and the x-axis are the true relation labels. Off-diagonal accuracy values show misclassifications for specific relations. }
    \label{fig:confusion_matrix_cnn_bilstm}
\end{figure*}

\subsection{Performance vs. Parameters Tradeoff}
\label{sec:tradeoff}
For the dataset, a combination of convolution and memory networks performs better for the relation classification task. The number of trainable parameters for CNN+LSTM is almost ten times that of the baseline model. Although the cost of training increases, this increment in the number of parameters leads to more information being learned by the deep learning model, which results in better performance over the baseline. Moreover, the CNN+BiLSTM outperforms the CNN+LSTM model as it holds the capacity to learn more information from the data due to more trainable parameters in the BiLSTM (10,000 parameters more than CNN+LSTM model). In addition, BiLSTM equips the model with the capability of learning context in both forward and reverse order. In fact, when we train models by increasing the number of parameters, the classification accuracy increases. However, the F1 score does not follow a similar trend. Our model has a comparable F1 score which should be sufficient for relation extractions, although the baseline model performs better in terms of F1 score. \par
As model complexity increases, so do the resources required for training the model. Compared to the baseline model, which has only 4,559 trainable parameters, our proposed model has a much higher number of parameters, significantly increasing training time. 
Although we do not investigate avenues of model interpretability in this work, models with more parameters generally tend to be less interpretable than models with fewer parameters. These factors should be considered when designing models for any task. Keeping this in mind, we used a max pooling layer after the CNN model to reduce the number of trainable parameters compared to the BiLSTM model without significant loss in generalizable performance.
As the CNN+LSTM/BiLSTM model has a higher performance, this will directly translate into more relations being accurately added to an existing knowledge graph.
Our model also converges faster than the baseline model (outperforming the previous model in terms of accuracy in about five epochs). This performance increase is likely due to the complexity of the model and more trainable parameters.\par

From the ablation study in section \ref{sec:ablation_study}, we observe that using just the BiLSTM model leads to performance gain over the CNN+BiLSTM model. However, the slight performance gain of 0.39\% points in accuracy and 1.89\% points in F1 score comes with the cost of a significant increase in the number of trainable parameters (36,472 more parameters than CNN+BiLSTM). This BiLSTM-only model leads to higher training time and a less interpretable architecture. Therefore, considering the computing cost and performance trade-off, we advocate for the CNN+BiLSTM for extracting relations from tabular data as a balance between the two extremes.\par

Fine-tuning BERT may also be beneficial for our task as fine-tuning approaches for language models have been shown to benefit the task at hand \cite{xue2019fine, Su2022, Liu2021}. However, fine-tuning can be extremely computationally extensive and may be impractical for scenarios where time is of importance. Moreover, fine-tuning BERT results in an increase in the number of trainable parameters, thus increasing the complexity of the model. Although beneficial for relation extraction, we used the embedddings from the pre-trained model in the interest of training and computation time.

\subsection{Difficult Relations}
\label{sec:difficult_relations}
We also wanted to investigate our model's ability to distinguish between difficult relations. We show a confusion matrix in Figure \ref{fig:confusion_matrix_cnn_bilstm} that depicts the accuracy of our proposed model for all the relation classes (we chose the model for the best performing seed value). Relations such as \texttt{director-film}, \texttt{actor-film}, \texttt{writer-film}, and \texttt{producer-film} are some of the most confusing examples for the model. This may be due to the fact that such relations are very similar to each other and is thus difficult for the model to distinguish one from the other. One may choose to provide extra information from the Wikipedia article or the table to the model for better understanding of the relations. More research is required to explore this idea.\par
As model complexity increases, so does the performance leading to better ability to distinguish between relations. However, this may not directly translate to high classification accuracy for difficult relations. A worthwhile direction to explore would be to design intelligent model training strategies that focus specifically on difficult relations without compromising performance on the rest of the classes.

\section{Conclusion and Future Work}
\label{sec:conclusion}

In this work, we proposed a neural method that uses a combination of convolution and memory networks to extract relations from Wikipedia tables, which we evaluate on a benchmark dataset. We also showed that combining convolution and max pooling helps to learn more about the data without a significant increase in the number of training parameters. We analyze our results and discuss the trade-off between the number of training parameters and model performance. Finally, we show how our model performs on relations that are deemed to be difficult to distinguish between and suggest some possible improvements for such cases.
We also conducted an ablation study to show the usefulness of the CNN layer. An extension of the ablation approach would be to remove certain input fields, like table cell values, headers, and captions, to evaluate model performance. An impactful idea in the space of relation extraction is the usage of the attention mechanism. Using the attention mechanism to identify tokens in the input that better represent a relation is a promising approach that may significantly improve tabular relation extraction. \par
We also highlight the trade-offs between parameters and the performance of the model as a first step toward probing relation extraction models. As neural network models become larger, it becomes even more crucial to provide explanations about the inner workings of the model. \par As neural network models grow larger with more training parameters, interpretability becomes crucial. In the future, we want to use sophisticated tools such as LIME \cite{ribeiro2016should} and SHAP \cite{lundberg2017unified} to explain how complex relation extraction models \textit{understand} the input to classify them into correct categories.

\bibliography{anthology,custom}






\end{document}